\documentclass[runningheads]{llncs}

\usepackage{multirow}
\usepackage{subfiles}
\usepackage{hyperref}
\usepackage{subcaption}
\usepackage{cite}
\usepackage{colortbl}
\usepackage{booktabs}
\usepackage{multirow}
\usepackage[table]{xcolor}
\definecolor{maroon}{cmyk}{0,0.87,0.68,0.32}
\usepackage{graphicx}
\usepackage[misc]{ifsym}
\usepackage{bbding}
\usepackage{color}

\begin{document}

\title{An Ensemble Method to Automatically Grade Diabetic Retinopathy with Optical Coherence Tomography Angiography Images}

\titlerunning{Ensemble Method for Diabetic Retinopathy Grading}

\author{Yuhan Zheng\inst{1}$^($\Envelope$^)$\orcidID{0000-0002-9762-6530}\and
Fuping Wu\inst{2} \orcidID{0000-0001-7179-4766} \and
Bart\l omiej W. Papie\.{z}\inst{2}$^($\Envelope$^)$\orcidID{0000-0002-8432-2511}}

\authorrunning{Y. Zheng et al.}

\institute{Department of Engineering, University of Oxford, Oxford, UK\\
\email{yuhan.zheng@spc.ox.ac.uk}
\and
Big Data Institute, University of Oxford, Oxford, UK\\
\email{fuping.wu@ndph.ox.ac.uk}, \email{bartlomiej.papiez@bdi.ox.ac.uk}
}

\maketitle

\begin{abstract}
Diabetic retinopathy (DR) is a complication of diabetes, and one of the major causes of vision impairment in the global population. 
As the early-stage manifestation of DR is usually very mild and hard to detect, an accurate diagnosis via eye-screening is clinically important to prevent vision loss at later stages. 
In this work, we propose an ensemble method to automatically grade DR using ultra-wide optical coherence tomography angiography (UW-OCTA) images available from Diabetic Retinopathy Analysis Challenge (DRAC) 2022. 
First, we adopt the state-of-the-art classification networks, i.e., ResNet, DenseNet, EfficientNet, and VGG, and train them to grade UW-OCTA images with different splits of the available dataset.
Ultimately, we obtain 25 models, of which, the top 16 models are selected and ensembled to generate the final predictions.
During the training process, we also investigate the multi-task learning strategy, and add an auxiliary classification task, the Image Quality Assessment, to improve the model performance. 
Our final ensemble model achieved a quadratic weighted kappa (QWK) of 0.9346 and an Area Under Curve (AUC) of 0.9766 on the internal testing dataset, and the QWK of 0.839 and the AUC of 0.8978 on the DRAC challenge testing dataset. 
\end{abstract}

\section{Introduction}
Diabetic retinopathy (DR) is the most common complication of diabetes and remains a leading cause of visual loss in working-age populations\cite{ref_9}. 
The current diagnosis pathway relies on the early detection of microvascular lesions\cite{ref_10}, such as microaneurysms, hemorrhages, hard exudates, non-perfusion and neovascularization. 
Based on its clinical manifestations, DR can be divided into two stages, i.e., non-proliferative DR (NPDR) and proliferative DR (PDR), corresponding to early and more advanced stages of DR, respectively. NPDR tends to have no symptoms, 
and it could take several years \cite{ref_11} to deteriorate into PDR, leading to severe vision impairment. 
Therefore, detection of early manifestation of DR is essential to accurate diagnosis and treatment monitoring. 

Optical coherence tomography angiography (OCTA), as a non-invasive imaging technique providing depth-resolved images for retinal vascular structure\cite{ref_12}, has become an effective imaging modality for DR diagnosis. 
However, manually identifying subtle changes on eye images is difficult and time-consuming.
Hence, computer-aided diagnosis of DR using OCTA images has attracted interest from researchers, and many computational approaches have been proposed in recent years, including traditional machine learning approaches such as decision tree and support vector machine \cite{ref_13, ref_14, ref_15}, and convolutional neural networks (CNNs) \cite{ref_16, ref_17, ref_18, ref_19, add_1, add_2, add_3}. 
Previous works can be mainly divided into two categories: (1) segmenting different types of lesions related to DR, and (2) classifying/ grading DR. However, identifying the best performing method requires the standardized datasets. 
Therefore, to foster the development of image analysis and machine learning techniques in clinical diagnosis of DR, and to address the lack of standardized datasets to make a fair comparison between the developed methods, the Diabetic Retinopathy Analysis Challenge (DRAC) is organized in conjunction with the Medical Image Computing and Computer Assisted Intervention Society (MICCAI) 2022.

In this work, we propose a deep learning model to automatically grade DR from ultra-wide (UW)-OCTA images into three classes, i.e., Normal, NPDR and PDR, with the aim to reduce the burden on ophthalmologists as well as providing a more robust tool to diagnose DR. 
We investigate four state-of-the-art CNNs for classification with different augmentation techniques, and an ensemble method is employed to generate final DR grading. 
We also utilize multi-task learning with the help of an auxiliary task i.e. Image Quality Assessment during the training process to further boost the performances of the main task, i.e., DR grading.

The remainder of this paper is organized as follows.
Section \ref{section dataset splitting} introduces the dataset and the preprocessing steps. 
The basic models used for DR grading, the ensemble strategy, and multi-task techniques are described in Section \ref{section baseline networks},  \ref{section model ensemble}, and \ref{section multi-task learning}, respectively.
Section \ref{exps} describes the experimental set-up and the results for ablation and evaluation.
Finally, the discussion and conclusion can be found in Section \ref{section discussion and conclusion}. 
The implementation of our method has been released via our GitHub repository.\footnote{https://github.com/Yuhanhani/DR-Grading-DRIMGA-.git}

\section{Methods}

\subsection{Dataset}
\label{section dataset splitting}

DRAC2022 includes three tasks (and the associated dataset) as follows.
Task 1: Segmentation Task provides 109 training images containing three types of lesions - intraretinal microvascular abnormalities, nonperfusion areas and neovascularization, and 65 testing images. 
Task 2: Image Quality Assessment contains 665 training images of three classes - Poor, Good and Excellent quality, and 438 testing images. 
Task 3: Diabetic Retinopathy Grading contains 611 images that is a subset of the previous 665 images (Task 2), and they are divided into three classes - Normal, NPDR and PDR. 
For challenge testing, 386 testing images are provided for DR grading task, however no expert annotations were available to the participants.

All of the images are grayscale images with the size of 1024 $\times$ 1024 pixels. The instrument used in this challenge is a swept-source (SS)-OCTA system (VG200D, SVision Imaging, Ltd., Luoyang, Henan, China), works near 1050 nm and features a combination of industry-leading specifications including ultrafast scan speed of 200,000 AScans per second\cite{ref_20}.

\begin{figure}[h]
\centering
\includegraphics[width=\textwidth,height=0.16\textheight]{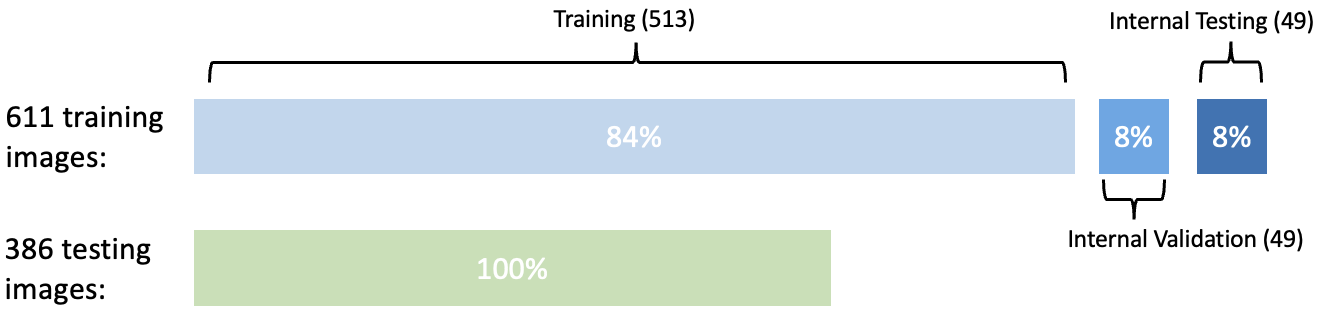}
\caption{The splitting strategy on the released training dataset. All of the 611 images were split into three sets: 513 images for training, 49 images for internal validation and 49 images for internal testing.
For testing, 386 testing images are provided for DR grading task, but with no expert annotations available to the participants.
}
\label{fig:figure1}
\end{figure}

Our models were mainly developed on DR grading dataset (Task 3). To develop and optimize models, we split the 611 training images into three subsets. 
Fig.\ref{fig:figure1} illustrates the way they were split, where 84\%, 8\%, 8\% form training, internal validation, and internal testing dataset respectively. 
The internal validation dataset was used to tune hyperparameters, select optimized models as well as for early stopping to prevent over-fitting. 
The testing dataset was used as a final sanity check before uploading the models to the challenge competition as well as for comparison between different networks, augmentations, and ensemble strategies presented below.
It is also worth mentioning that the overall class distributions of the 611 images are highly imbalanced, with the three classes occupying 53.8\% for Normal, 34.7\% for NPDR, and 11.5\% for PDR, respectively. 
Therefore, to generate our internal validation and testing dataset, we performed a stratified sampling\cite{ref_1}. Each of the three subsets was guaranteed to follow the same distribution as the overall distribution, so the internal validation and testing dataset can better represent the entire population that was being studied.

\subsection{Baseline Networks}
\label{section baseline networks}
We adopted four state-of-the-art CNN architectures, including ResNet\cite{ref_2}, \\
DenseNet\cite{ref_3}, EfficientNet\cite{ref_5} and VGG\cite{ref_4}. 

\subsubsection{ResNet.}
This is one of the most commonly used networks that achieves high performance on classification tasks.
Many deep learning methods for retinal images classification are developed using ResNet as a backbone\cite{ref_22, ref_23, ref_24}.
ResNet is a deep residual learning framework, which addresses vanishing-gradient and degradation problem, and eases the training of deeper networks by using residual mapping \cite{ref_2}.

\subsubsection{DenseNet.}
DenseNet was the another network used for our DR grading, as it is reported to have better feature use efficiency with fewer parameters than ResNet. It utilizes dense blocks, in which each layer is connected to every other layer in a feed-forward manner. It also alleviates the problem of vanishing-gradient, and increases feature reuse and reduces the number of parameters required\cite{ref_3}.

\subsubsection{EfficientNet.}
EfficientNet is a set of models obtained by uniformly scaling all dimensions of depth, width, and resolution using compound coefficients. EfficientNet is shown to achieve better performances on ImageNet, and also to transfer well on other datasets\cite{ref_5}. 
It was also employed on our dataset and brought the benefits of small model size with fast computation speed.

\subsubsection{VGG.}
We finally tried VGG, which is an early CNN architecture that forms the basis of object recognition models. By utilizing small convolutional filters and increasing the depth of CNNs to 16-19 weight layers, it shows outstanding performances on ImageNet\cite{ref_4}. It is also used on OCTA imaging showing good performances\cite{ref_21}. However, its downsides are the large model size and long computation time. 

\subsection{Model Ensemble}
\label{section model ensemble}

Following the splitting manner introduced in Section \ref{section dataset splitting}, we additionally split the entire dataset randomly three times (A, B, C), such that each time we had different training and internal validation set, while the internal testing set was kept unchanged. 
Then for each split, all models were trained, and optimized to select the best performing models. 
The Table \ref{tab:table1} summarizes the selection of 16 models used for ensemble method. 

\begin{table}[h]
    
    \caption{Summary of all the selected models for given data split (A, B, and C). The number of models selected for each architecture resulting in 16 models in total used for ensemble method.}
    \vspace{3mm}
    \centering
    \begin{tabular}{c|c|c|c|c|c}
        \hline 
        split & ResNet34 & ResNet152 & DenseNet121 & VGG19\_BN & EfficientNet\_b0\\
        \hline
        A & one model & one model & None & one model & one model \\
        \hline
        B & None & one model & two models & two models & two models \\
        \hline
        C & two models & None & one model & one model & one model \\
        \hline
    \end{tabular}
    \label{tab:table1}
\end{table}

\begin{figure}[h]
\centering
\includegraphics[width=0.8\textwidth,height=0.238\textheight]{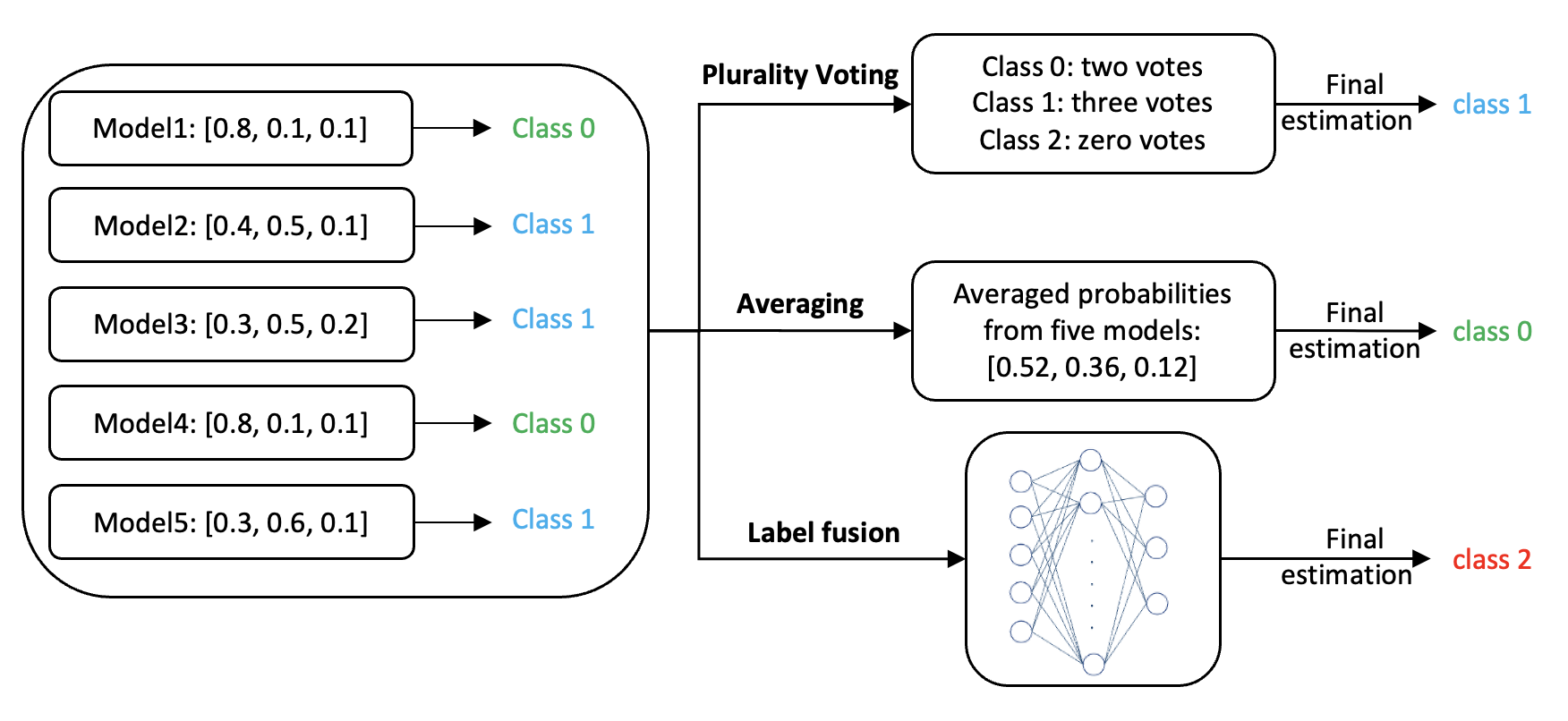}
\caption{Diagram shows the example of each ensemble strategy using five models for three-class classification. The predicted probabilities for each class are shown in square brackets.}
\label{fig:figure6}
\end{figure}

\begin{figure}[h]
\centering
\includegraphics[width=0.8\textwidth,height=0.23\textheight]{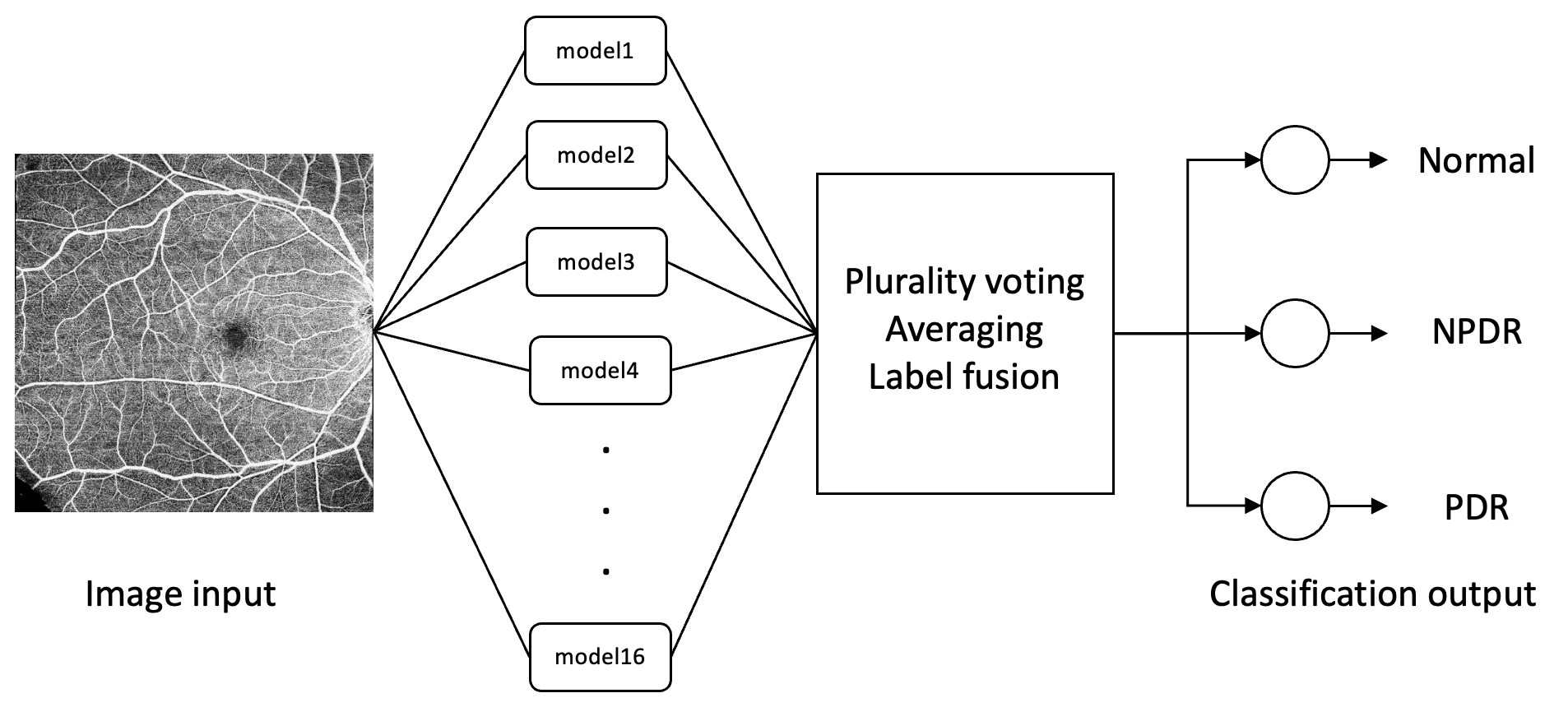}
\caption{Schematic diagram of model ensemble strategies where an image is input and the final DR grading is generated by aggregating 16 models.}
\label{fig:figure3}
\end{figure}

Model ensemble is a machine learning technique to combine outputs produced by multiple models in the prediction process, and it overcomes drawbacks associated with a single estimator such as high variance, noise, and bias\cite{ref_6}.
In this work, three ensemble strategies were adopted. 
The first one was plurality voting, which takes the class that receives the highest number of votes as the final prediction\cite{ref_7}. 
The second was averaging, which outputs the final probabilities as the unweighted average of probabilities estimated by each model. 
The third technique was label fusion with a three-layer simple neural network, which was trained with inputs as the predictions from the 16 models, and learns to assign appropriate weights to each individual model. An example illustrating these three strategies using five models and three classes can be found in Fig.\ref{fig:figure6}. 
It can be seen that the final prediction output can be completely different depending on ensemble strategy used. 
Fig.\ref{fig:figure3} shows our ensemble method for DR grading.

\subsection{Multi-task Learning}
\label{section multi-task learning}

\begin{figure}[h]
\centering
\includegraphics[width=0.6\textwidth,height=0.185\textheight]{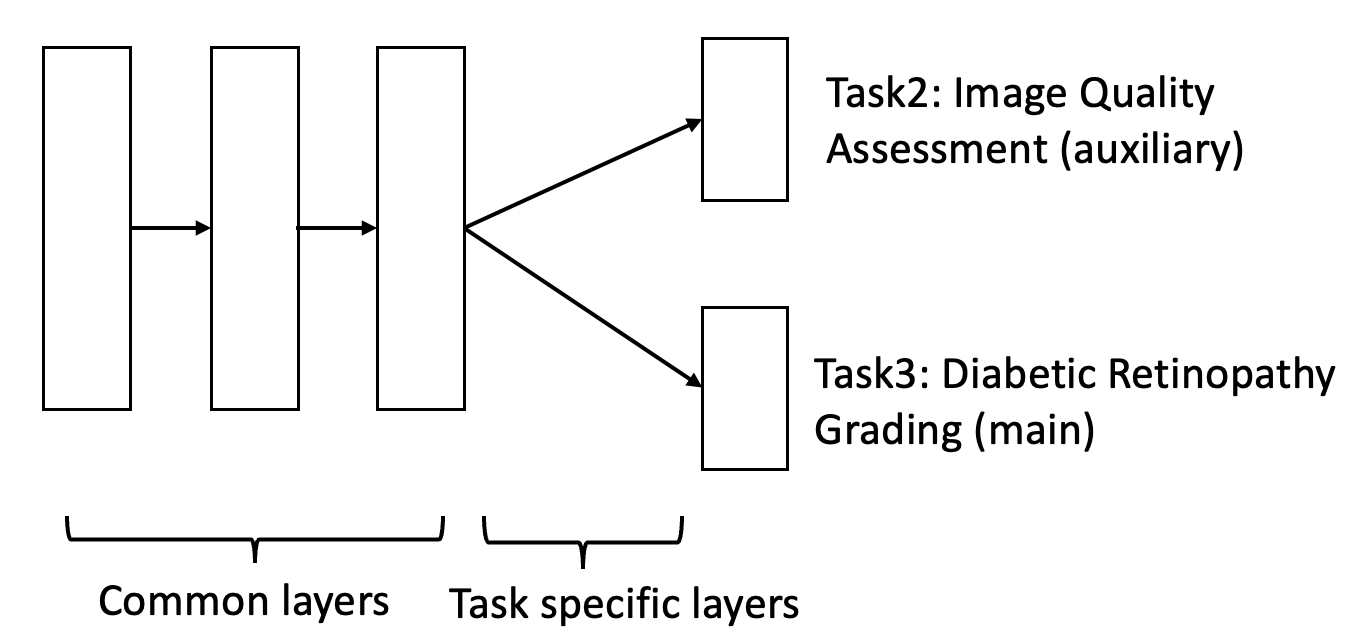}
\caption{Schematic diagram of proposed multi-task learning.}
\label{fig:figure4}
\end{figure}

We used multi-task learning throughout our network training process, in order to take advantage of complementary information from different tasks, and to improve the robustness of the model. 
We simultaneously trained model for Image Quality Assessment and Diabetic Retinopathy Grading via hard parameter sharing \cite{ref_8}, as shown in Fig.\ref{fig:figure4}. 
Particularly, Image Quality Assessment Task was treated as an auxiliary task, with the aim to obtain a more generalized model thereby improving model performances for the main task, i.e., DR grading. The loss function was defined as follows.
\begin{equation}
    Loss_{total} = Loss_{task3} + \lambda\cdot Loss_{task2}
    \label{eq:eq1}
\end{equation}

\noindent where $\lambda$ is a hyperparameter. In our experiment, $\lambda = 0, 0.01, 0.1, 1$ was the set of values that were tested to improve the performances of DR grading task in the different splits (A,B,C).

\section{Experiments}\label{exps}

\subsection{Experimental setting}
\label{section experimental setting}
\subsubsection{Image Preprocessing.}

\begin{figure}[h]
\centering

\begin{subfigure}{0.24\textwidth}
\centering
\includegraphics[width=\textwidth,height=0.14\textheight]{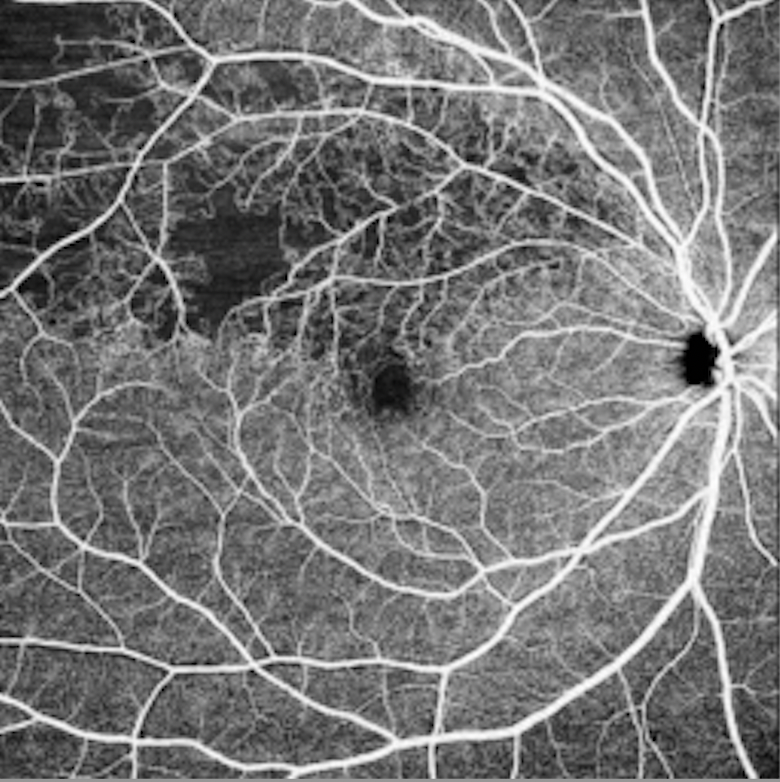}
\caption{original image}
\label{fig:image2_1}
\end{subfigure}
\begin{subfigure}{0.24\textwidth}
\centering
\includegraphics[width=\textwidth,height=0.14\textheight]{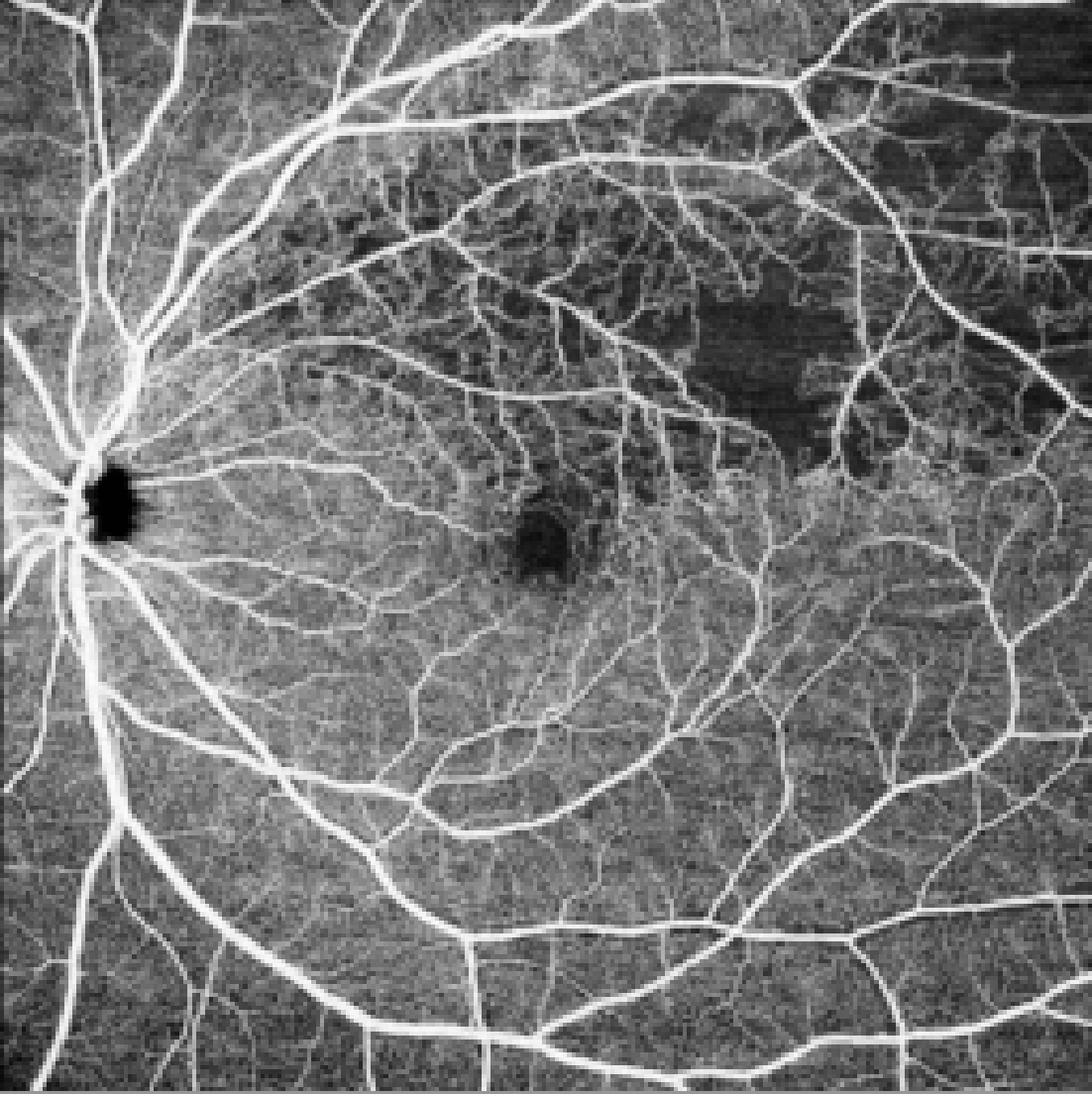}
\caption{random flip}
\label{fig:image2_2}
\end{subfigure}
\begin{subfigure}{0.24\textwidth}
\centering
\includegraphics[width=\textwidth,height=0.14\textheight]{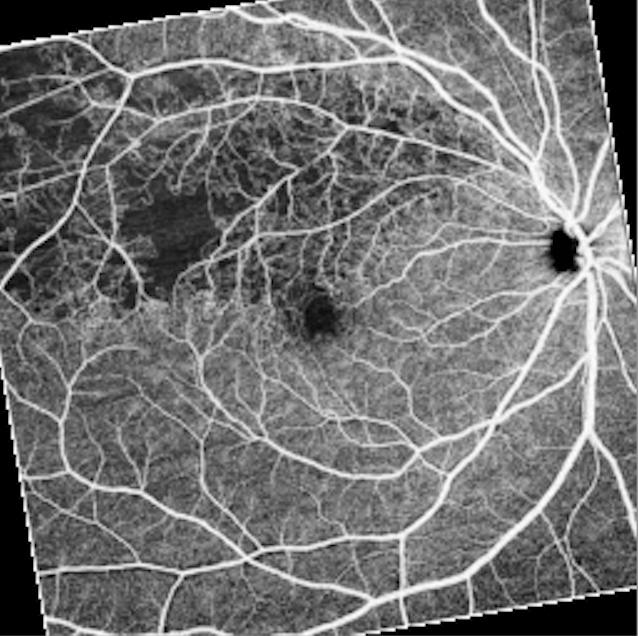}
\caption{random rotation}
\label{fig:image2_3}
\end{subfigure}
\begin{subfigure}{0.24\textwidth}
\centering
\includegraphics[width=\textwidth,height=0.14\textheight]{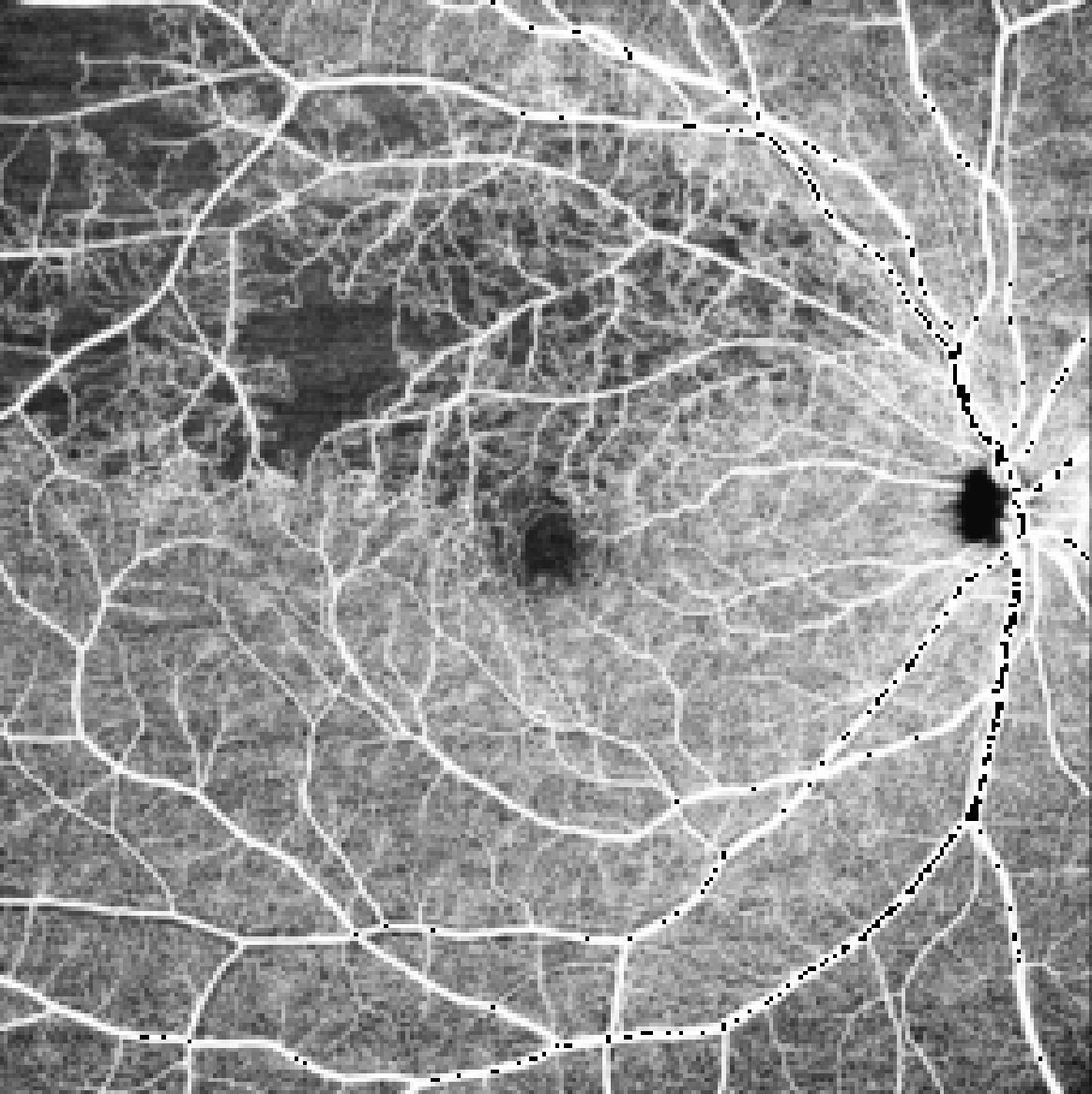}
\caption{color jitter}
\label{fig:image2_4}
\end{subfigure}

\caption{Examples of (a) the original UW-OCTA image, (b) after random flip, (c) random rotation, (d) and color jitter (contrast and brightness adjusted). The images shown here were resized to the size of 256 $\times$ 256 pixels.}
\label{fig:image2}
\end{figure}

For all experiments, the images were resized to either 256 $\times$ 256 or 512 $\times$ 512 pixels from its original size of 1024 $\times$ 1024 pixels. We further employed image augmentation techniques including random horizontal flip, random vertical flip, random rotation, and color jitter.
Fig. \ref{fig:image2} shows examples of UW-OCTA images with augmentation applied.

\subsubsection{Training Hyper-parameters}
All the models were pretrained on ImageNet Dataset\cite{ref_25}. 
Throughout the training process, we used the stochastic gradient descent (SGD) optimizer with a learning rate scheduler with exponential decay. The initial learning rate was set to 0.001 and the decay factor was set to 0.8 or 0.9. 
We trained the network using 20 epochs and saved the model that gave the best performance measured by a quadratic weighted kappa (QWK).
For $Loss_{task3}$, a weighted cross-entropy loss was used. We increased penalty mostly for NPDR class, as it was the most difficult class to predict for most of the cases. The weights were adjusted empirically. 
For $Loss_{task2}$, the weighted cross entropy loss was used with the weight values of 0.779, 0.146, and 0.075, based on the Task 2 class distributions. 
The batch size was 25.

\subsubsection{Evaluation Metrics}
\label{section evaluation metrics}
For evaluation, we used metrics provided by the challenge organizer i.e.  a Quadratic Weighted Kappa (QWK) and an Area Under Curve (AUC) of the receiver operating characteristic (ROC) curve. 
Specifically, One-vs-One (ovo) macro-AUC was used. The challenge ranking was based on the QWK, and if  QWK were the same, then ovo-macro-AUC was used as a secondary ranking metric.

\subsection{Evaluation of Image Preprocessing}
\label{section effect of image preprocessing}

Here we study how data augmentation i.e. color jitter and resizing influence the model performance. 
As shown in Table \ref{tab:table2a}, for all of the networks except DenseNet, the use of color jitter boosts the performance. Therefore, we incorporated color jitter into our data augmentation process. 
In addition, Table \ref{tab:table2b} shows that resizing images down to the size of 256 $\times$ 256 pixels did not lead to noticeable performance degradation compared to those with the original image size.
The use of lower image resolution made it possible to train with deeper networks and larger batch size. 
In contrast, with the size of 128 $\times$ 128 pixels, the performance of the tested models was consistently lower, which could be due to the loss of information related to the disease. 

\begin{table}[!htb]

\caption{Quadratic Weighted Kappa (QWK) for different models (a) with and without color jitter. (b) and with the varying size of the image, evaluated on the internal validation dataset (average across all splits).}
\vspace{3mm}
\begin{subtable}[c]{.5\linewidth}
\centering
\caption{results of applying color jitter}
\begin{tabular}{l|p{1.5cm}|p{1.5cm}}
\hline
{}&with&without\\
\hline
ResNet 34 & \textbf{0.7492} & 0.7310\\
\hline
DenseNet 121 & 0.7659 & \textbf{0.7850}\\
\hline
VGG19\_BN & \textbf{0.7969} & 0.7932\\
\hline
EfficientNet\_b0 & \textbf{0.7707} & 0.7695\\
\hline
\end{tabular}
\label{tab:table2a}
\end{subtable}
\begin{subtable}[c]{.5\linewidth}
\centering
\caption{results of resizing image}
\begin{tabular}{l|l|l|l|l}
\hline
Size & \textbf{128$^2$} & 256$^2$ & 512$^2$ & 1024$^2$\\
\hline
ResNet 34 & 0.5315 & 0.7492 & \textbf{0.7713} & 0.7561\\
\hline
DenseNet 121 & 0.5922 & 0.7659 & 0.7965 & \textbf{0.8569}\\
\hline
\end{tabular}
\label{tab:table2b}
\end{subtable}

\label{tab:table2}
\end{table}

\subsection{Evaluation of Model Ensemble}
\label{section effect of model ensemble}

Here we present the results of different ensemble methods.
Table \ref{tab:table3a} and \ref{tab:table3b} show the performances of the individual networks and the ensemble methods, evaluated on the internal testing dataset. 
For the individual model evaluation, VGG19\_BN obtained the best performances (the QWK of 0.8715 and the AUC of 0.9665) comparing to the other models.
For the ensemble method evaluation, label fusion produced the best result (the QWK of 0.9346 and the AUC of 0.9766) out of the three ensemble techniques. 
The use of the ensemble method boosted the result by about 6\% when compared to single VGG19\_BN.

\begin{table}[!htb]
\caption{The Quadratic Weighted Kappa (QWK) and the Area Under Curve (AUC)
of the individual networks used in this study, and three ensemble models, evaluated on the internal testing dataset (average across all splits).}
\vspace{3mm}

\begin{subtable}[c]{.5\linewidth}
\centering
\caption{Individual networks}
\begin{tabular}{p{2.5cm}|p{1.5cm}|p{1.5cm}}
\hline
Networks & QWK & AUC\\
\hline
ResNet 34 & 0.8663 & 0.9423\\
\hline
ResNet 152 & 0.8170 & 0.9350 \\
\hline
DenseNet 121 & 0.8487 & 0.9583\\
\hline
\textbf{VGG19\_BN} & \textbf{0.8715} & \textbf{0.9665}\\
\hline
EfficientNet\_b0 & 0.8270 & 0.9420\\
\hline
\end{tabular}
\label{tab:table3a}
\end{subtable}
\begin{subtable}[c]{.5\linewidth}
\centering
\caption{Ensemble models}
\begin{tabular}{p{2.8cm}|p{1.5cm}|p{1.5cm}}
\hline
Ensemble method & QWK & AUC\\
\hline
Plurality voting & 0.9131 & 0.9766\\
Averaging & 0.9131 & 0.9677\\
\textbf{Label fusion} & \textbf{0.9346} & \textbf{0.9766}\\
\hline
\end{tabular}
\label{tab:table3b}
\end{subtable} 

\label{tab:table3}
\end{table}

\subsection{Evaluation of Multi-task Learning}
Table \ref{tab:table4} shows the effect of multi-tasking of the three different ensemble strategies. 
It can be seen that single task outperformed multi-tasking in two cases when using our internal testing dataset.
However, it will be shown in Table \ref{tab:table5} in Section \ref{section final results} that multi-tasking boosted the performance by 2\% on the challenge testing dataset. 
The reason for this discrepancy could be the fact that our internal testing dataset had only 49 OCTA images, while there were 386 challenge testing images. 
Therefore, an increase of 2\% on the larger challenge testing dataset was considered far more significant to conclude that the multi-tasking can improve the overall performances.

\begin{table}[!htb]
\centering
\caption{The Quadratic Weighted Kappa (QWK) and the Area Under Curve (AUC) of single task and multi-tasking using the three ensemble methods, evaluated on the internal testing dataset.}
\vspace{3mm}

\begin{tabular}{l|l|p{1.1cm}|p{1.1cm}|l|l|p{1.1cm}|p{1.1cm}}
\hline
 & Strategies & QWK & AUC &  & Strategies & QWK & AUC \\
\hline
\multirow{3}{*}{\begin{tabular}{l}Single\\Task\end{tabular}} & Plurality voting & 0.9167 & 0.9710 & \multirow{3}{*}{\begin{tabular}{l}Multi- \\tasking\end{tabular}} & Plurality voting & 0.9131 & 0.9766 \\
& Averaging & 0.9351 & 0.9677 & & Averaging & 0.9131 & 0.9677 \\
& Label fusion & 0.9322 & 0.9677 & & Label fusion & 0.9346 & 0.9766 \\
\hline
\end{tabular}

\label{tab:table4}
\end{table}

\subsection{Final Challenge Evaluation}
\label{section final results}

The results of our four final submissions to the DRAC challenge testing dataset are shown in Table \ref{tab:table5}. 
It can be seen that the best QWK of 0.8390 and the AUC of 0.8978 were achieved using averaging with multi-tasking loss function.
It should be noticed that the presence of multi-task learning boosted the final testing result by about 2\%.
In addition, for ensemble methods, averaging produced the best result as opposed to the internal testing case where label fusion was the best. 

\begin{table}[!htb]
\centering
\caption{The Quadratic Weighted Kappa (QWK) and the Area Under Curve (AUC) for different training and ensemble strategies on the DRAC challenge testing dataset.}
\vspace{3mm}
\begin{tabular}{l|l|p{1.1cm}|p{1.1cm}|l|l|p{1.1cm}|p{1.1cm}}
\hline
 & Strategies & QWK & AUC &  & Strategies & QWK & AUC \\
\hline
\multirow{3}{*}{\begin{tabular}{l}Single\\Task\end{tabular}} & Plurality voting & 0.8160  & 0.9046 & \multirow{3}{*}{\begin{tabular}{l}Multi- \\tasking\end{tabular}} & Plurality voting & 0.8308 & 0.8998 \\
& Averaging & 0.8204 & 0.9062 & & \textbf{Averaging} & \textbf{0.8390} & \textbf{0.8978} \\
& Label fusion & 0.8089 & 0.8928 & & Label fusion & 0.8276 & 0.8922 \\
\hline
\end{tabular}

\label{tab:table5}
\end{table}

\section{Discussion and Conclusion}
\label{section discussion and conclusion}

In this work, we presented the results of our investigation for the best method to perform diabetic retinopathy grading automatically. 
We used the state-of-the-art classification networks i.e. ResNet, DenseNet, EfficientNet and VGG as our baseline networks. We selected 16 models with different data splitting strategies, and utilized the ensemble method to produce the final prediction. 
We showed that hyperparameter tuning, data augmentation, and the use of the multi-task learning, with the auxiliary task of Image Quality Assessment, boosted the main task performances. 
Our final submission to the DRAC challenge produced the QWK of 0.8390 and the AUC of 0.8978.

\begin{figure}[h]
\centering

\begin{subfigure}{0.3\textwidth}
\centering
\includegraphics[width=\textwidth,height=0.19\textheight]{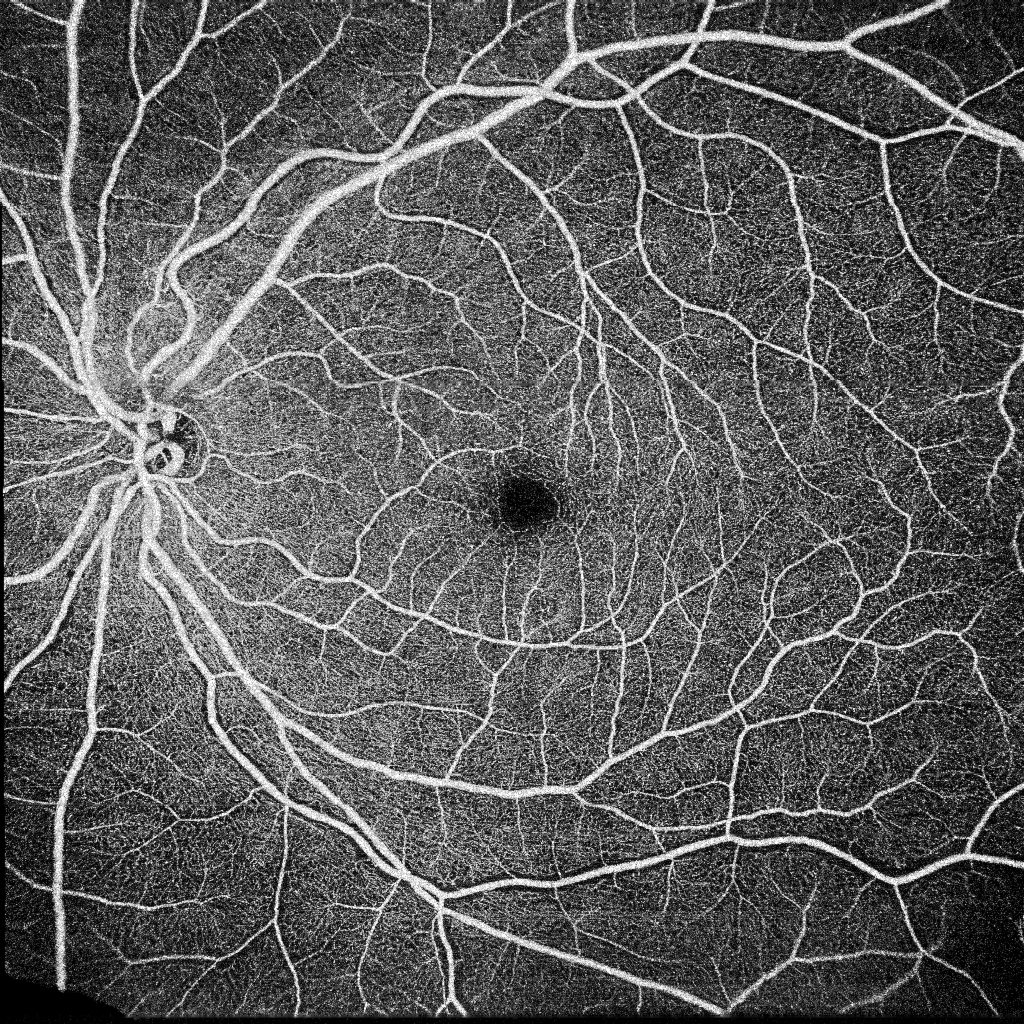}
\caption{true Normal}
\label{fig:image7_1}
\end{subfigure}
\begin{subfigure}{0.3\textwidth}
\centering
\includegraphics[width=\textwidth,height=0.19\textheight]{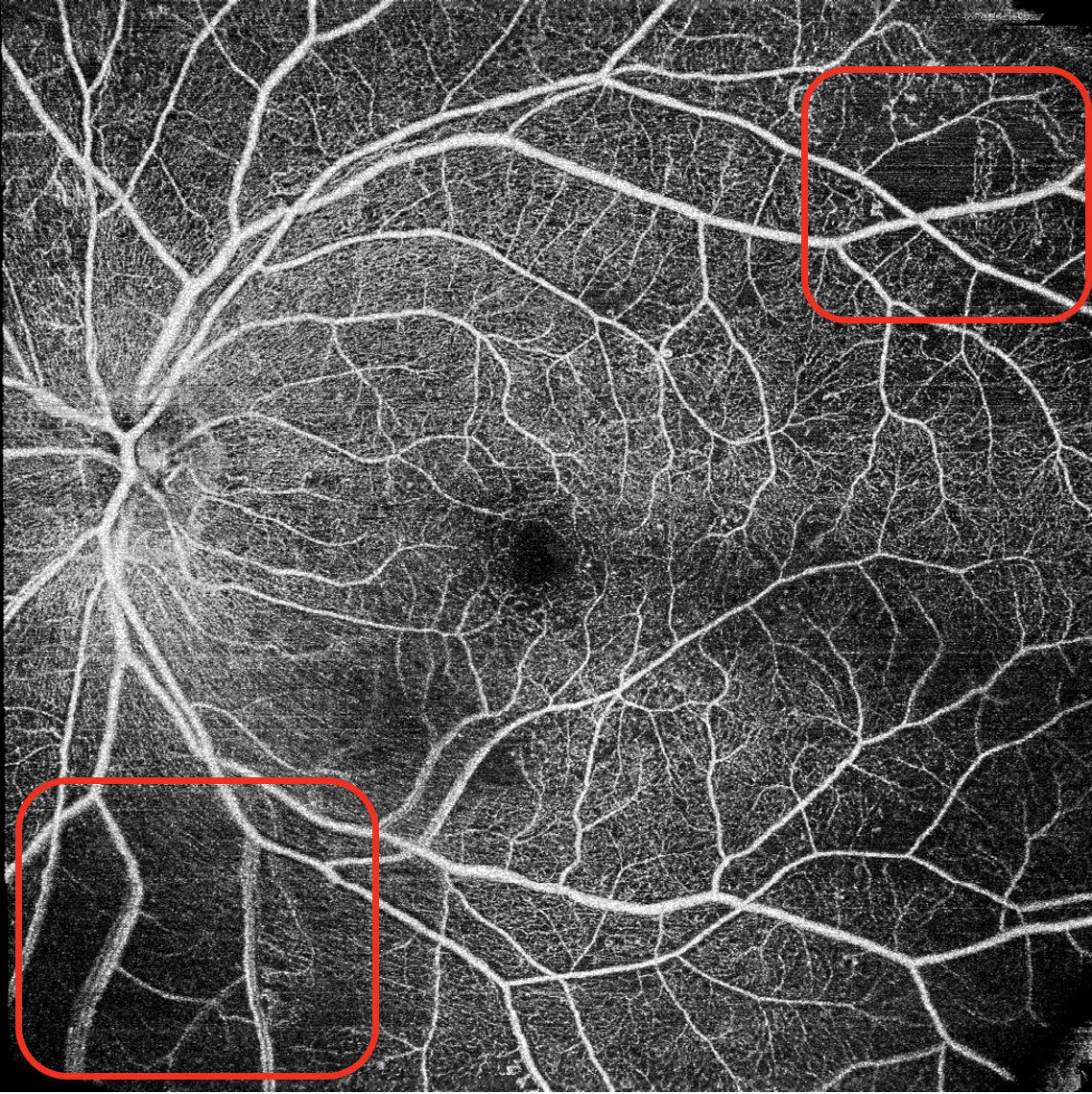}
\caption{true NPDR}
\label{fig:image7_2}
\end{subfigure}
\begin{subfigure}{0.3\textwidth}
\centering
\includegraphics[width=\textwidth,height=0.19\textheight]{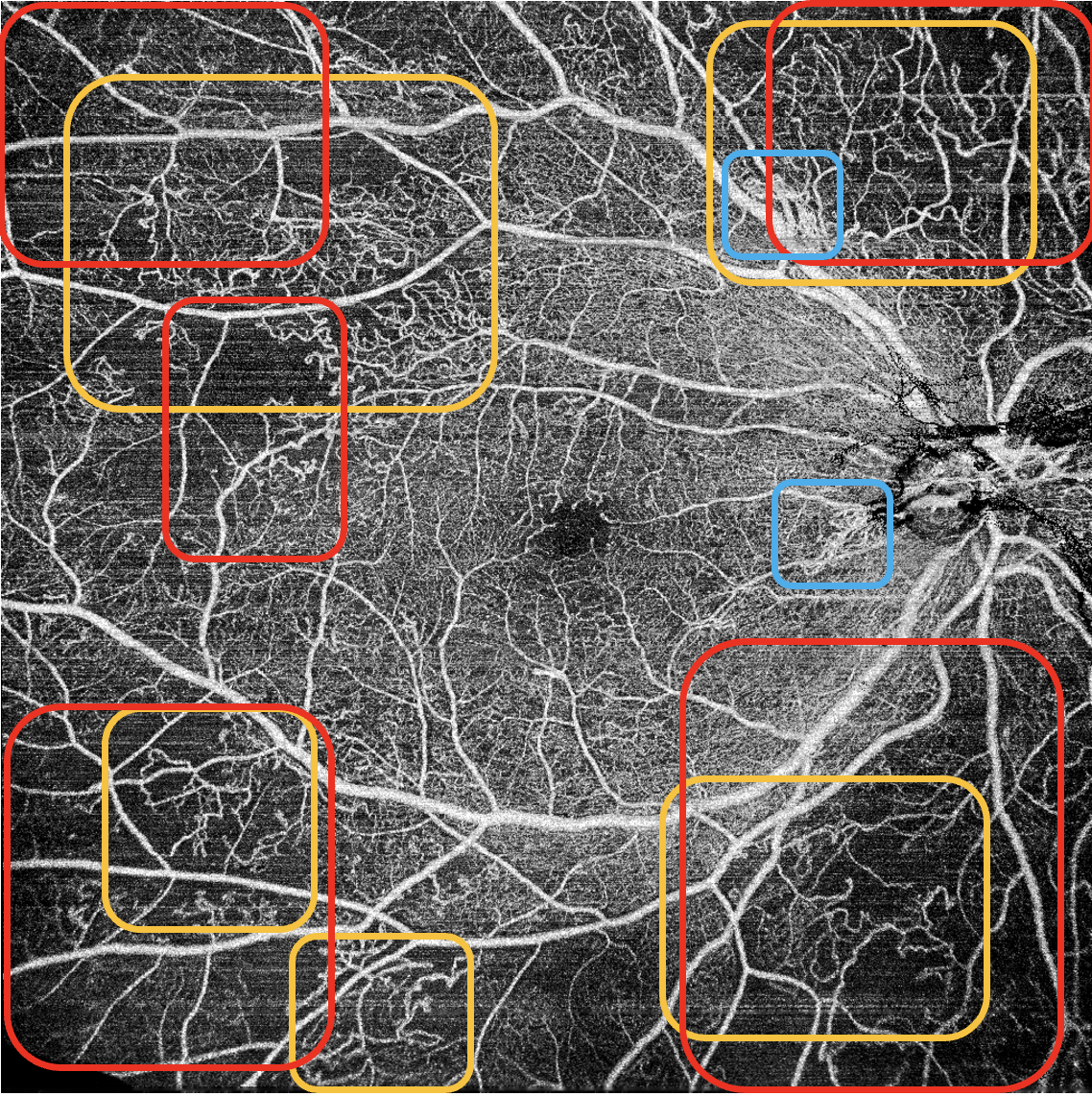}
\caption{true PDR}
\label{fig:image7_3}
\end{subfigure}
\begin{subfigure}{0.3\textwidth}
\centering
\includegraphics[width=\textwidth,height=0.19\textheight]{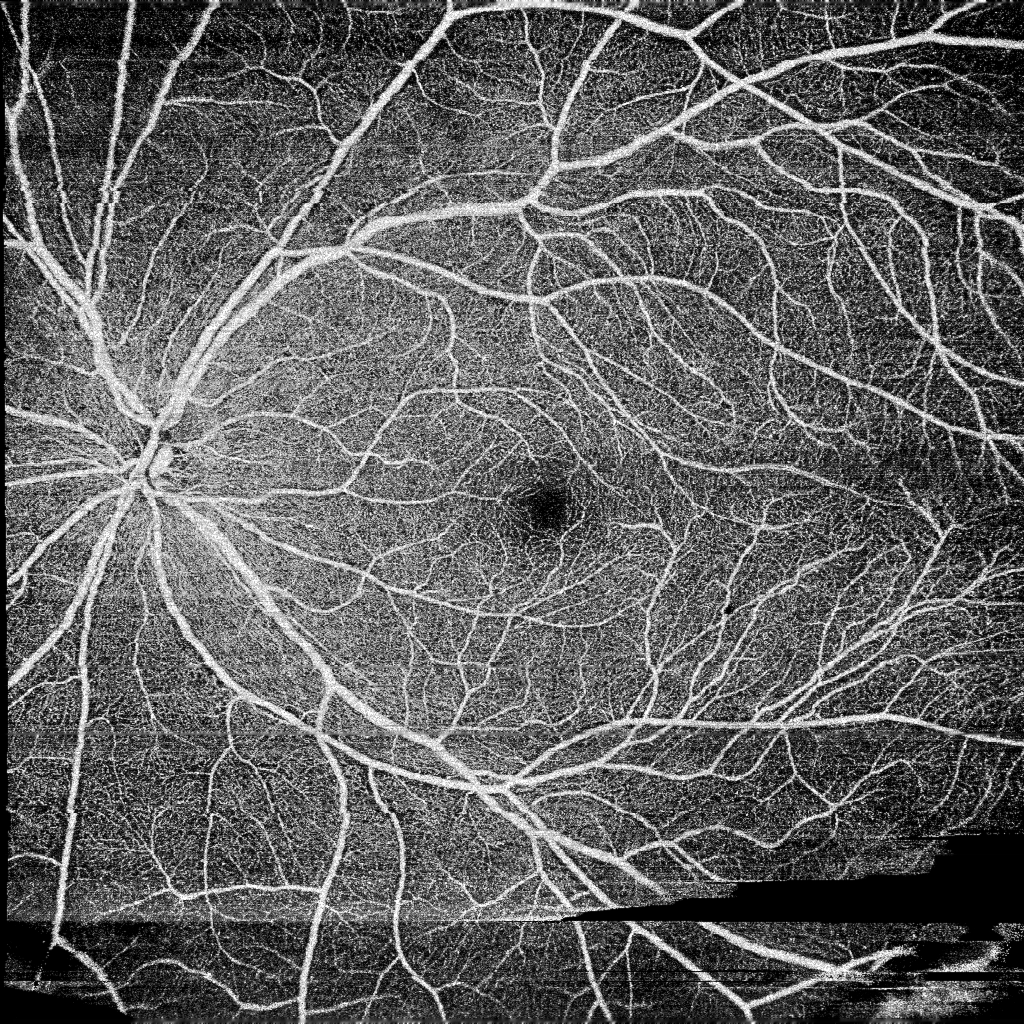}
\caption{false NPDR (Normal)}
\label{fig:image7_4}
\end{subfigure}
\begin{subfigure}{0.3\textwidth}
\centering
\includegraphics[width=\textwidth,height=0.19\textheight]{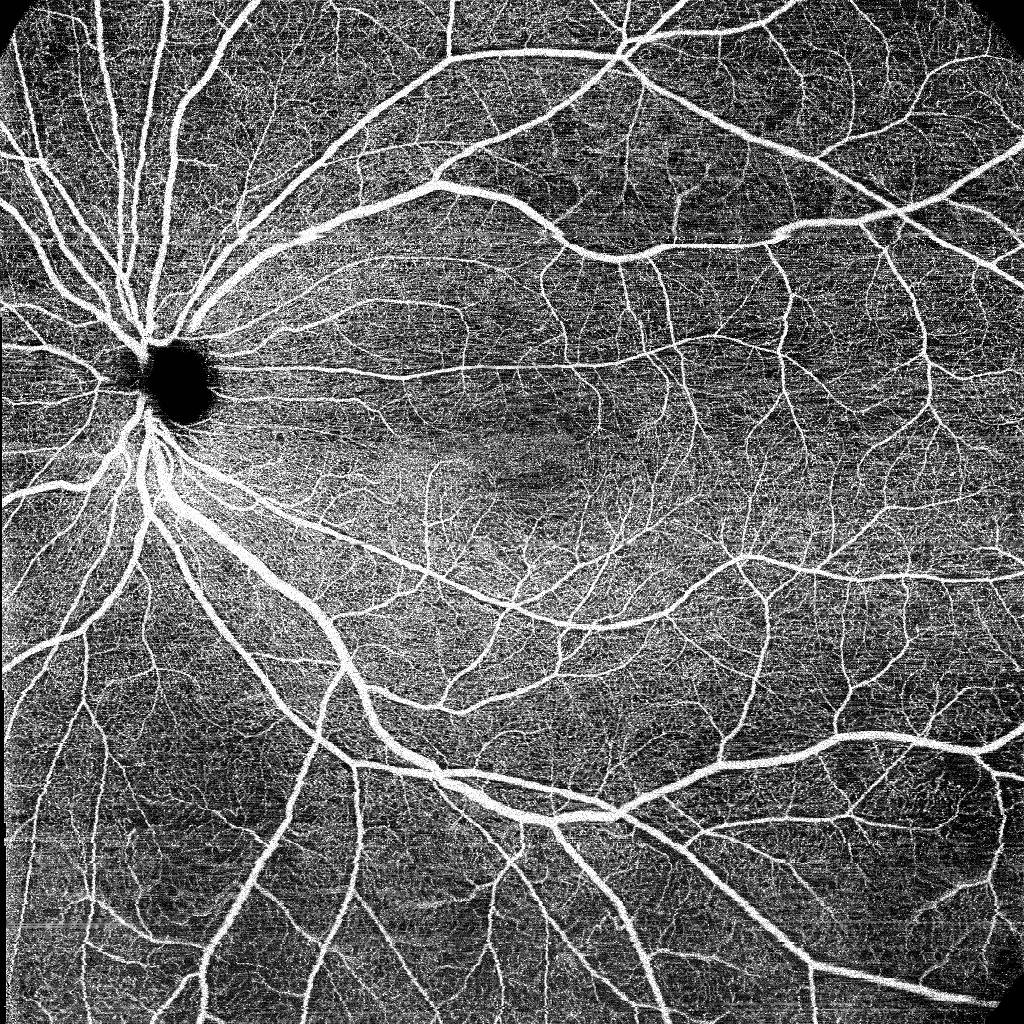}
\caption{false Normal (NPDR)}
\label{fig:image7_5}
\end{subfigure}
\begin{subfigure}{0.3\textwidth}
\centering
\includegraphics[width=\textwidth,height=0.19\textheight]{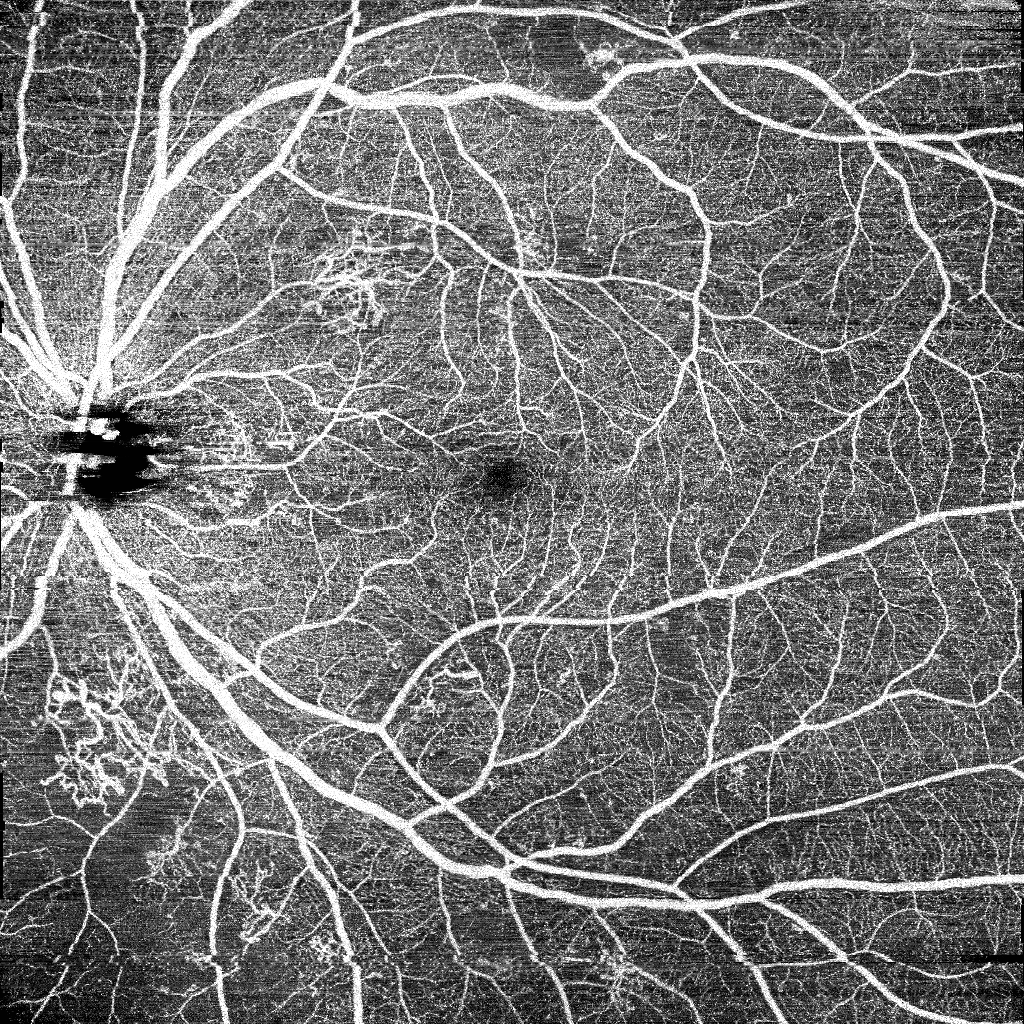}
\caption{false NPDR (PDR)}
\label{fig:image7_6}
\end{subfigure}

\caption{Examples of six images with labels predicted by our final ensemble model. Image (a),(b) and (c) are three correctly predicted images belonging to Normal, NPDR, PDR respectively. Image (d), (e) and (f) are three incorrectly predicted images with their true labels shown in the bracket. The colored boxes in Image (b) and (c) highlight different types of lesions with yellow, red and blue box indicating intraretinal microvascular abnormalities, nonperfusion areas and neovascularization respectively. They are drawn based on the ground truth provided for the Segmentation Task. 
}
\label{fig:image7}
\end{figure}

By comparing Table \ref{tab:table3} and Table \ref{tab:table4} it can be observed that the results on our internal testing dataset were always higher than on the challenge testing dataset. This might be due to the fact that, there were only 49 testing images in our internal dataset, which could present distribution bias against the final testing dataset,
and our models might have been over-fitted. 
Therefore, in the future, pretraining on other public OCTA datasets could be used. 
The three random splits we used, also may also produce optimistic results, hence a nested k-fold cross validation should be attempted to give more generalized results \cite{nested}. 
In addition, attention mechanism could be employed to pick up the most informative region \cite{Bourigault22}.
Furthermore, Fig.\ref{fig:image7} shows three correctly and three incorrectly predicted images by the final model. The three correctly predicted images are relatively standard Normal, NPDR and PDR images. Fig.\ref{fig:image7_1} is Normal and has no lesions. Fig.\ref{fig:image7_2} has some nonperfusion areas (red box) near the corner and is classified as NPDR. 
There are some surrounding intraretinal microvascular abnormalities (yellow box), nonperfusion areas (red box) and a small proportion of neovascularization (blue box) on Fig.\ref{fig:image7_3}, which conclude being PDR case. 
Our model, however, predicted incorrectly for the cases such as poorer image quality or non-obvious lesion appearances. 
This issue could be alleviated by e.g. adopting active learning strategy, which determines hard samples to be used for retraining.
Additionally, to regularize the model we chose Image Quality Assessment as the auxiliary task, which is intuitively less correlated to the main task than Task 1, i.e., Segmentation of Diabetic Retinopathy Lesions. However, for deep neural networks, inter-task relationships \cite{multi_2} could be difficult to be defined or measured. Recently, many researches have been done in this area to rigorously choose the optimal task grouping via proposing measurements for task relationship, such as inter-task affinity\cite{multi_1}, i.e. the effect to which one task’s gradient would affect another task’s loss. Hence, we could study in the future how the three challenge tasks could benefit each other by exploring their relationships in multi-task learning framework. Generally, our main task benefited from implicit data augmentation and regularization\cite{multi_3} of multi-tasking, which made it more robust against random noises, and managed to learn more general features, thus yielding a better overall performance.

\section*{Acknowledgment}
The authors acknowledge the DRAC2022 challenge for available UW-OCTA images for this study \cite{drac2022}.
The authors would like to thank Dr. Le Zhang from University of Oxford  for helpful comments on our manuscript. 


\bibliographystyle{splncs04}
\bibliography{mybibliography}

\end{document}